\documentclass[sigconf]{acmart}

\usepackage{soul}
\usepackage{url}
\usepackage{balance}
\usepackage{graphicx}
\usepackage{amsmath}
\usepackage{booktabs}
\usepackage{algorithm}
\usepackage[switch]{lineno}
\usepackage{algpseudocode} 
\usepackage{multirow}
\usepackage{threeparttable} 
\usepackage{makecell}
\usepackage{color}
\usepackage{balance}
\def\etal{\textit{et al.}}
\def\ie{\textit{i.e.}}
\def\eg{\textit{e.g.}}
\newcommand{\mhy}[1]{\textcolor[rgb]{0,0,0}{#1}}

\copyrightyear{2023}
\acmYear{2023}
\setcopyright{acmlicensed}
\acmConference[MM '23] {Proceedings of the 31st ACM International Conference on Multimedia}{October 29--November 3, 2023}{Ottawa, ON, Canada}
\acmBooktitle{Proceedings of the 31st ACM International Conference on Multimedia (MM '23), October 29--November 3, 2023, Ottawa, ON, Canada}
\acmPrice{15.00}
\acmISBN{979-8-4007-0108-5/23/10}
\acmDOI{10.1145/3581783.3612147}

\begin{document}

\title{Event-Enhanced Multi-Modal Spiking Neural Network for Dynamic Obstacle Avoidance}

\author{Yang Wang}
\affiliation{%
 \institution{Dalian University of Technology}
 \city{Dalian}
 \country{China}}
 \email{yangwang06@mail.dlut.edu.cn}

 \author{Bo Dong}
\affiliation{%
 \institution{Princeton University}
 \city{Princeton}
 \country{USA}}
 \email{bo.dong@princeton.edu}

 \author{Yuji Zhang}
\affiliation{%
 \institution{Dalian University of Technology}
 \city{Dalian}
 \country{China}}
 \email{yujizhang@mail.dlut.edu.cn}

  \author{Yunduo Zhou}
\affiliation{%
 \institution{Dalian University of Technology}
 \city{Dalian}
 \country{China}}
 \email{1227627334@mail.dlut.edu.cn}

 \author{Haiyang Mei}
\affiliation{%
 \institution{Dalian University of Technology \& National University of Singapore}
 \city{Dalian}
 \country{China}}
 \email{haiyang.mei@outlook.com}

 \author{Ziqi Wei}
\affiliation{%
 \institution{Institute of Automation, Chinese Academy of Sciences \& Dalian University of Technology}
 \city{Beijing}
 \country{China}}
 \email{ziqi.wei@ia.ac.cn}

 \author{Xin Yang}
 \authornote{Corresponding author.}
\affiliation{%
 \institution{Dalian University of Technology}
 \city{Dalian}
 \country{China}}
 \email{xinyang@dlut.edu.cn}

\renewcommand{\shortauthors}{Yang Wang et al.}

\begin{abstract}
Autonomous obstacle avoidance is of vital importance for an intelligent agent such as a mobile robot to navigate in its environment. Existing state-of-the-art methods train a spiking neural network (SNN) with deep reinforcement learning (DRL) to achieve energy-efficient and fast inference speed in complex/unknown scenes. These methods typically assume that the environment is static while the obstacles in real-world scenes are often \textbf{\textit{dynamic}}. The movement of obstacles increases the complexity of the environment and poses a great challenge to the existing methods. In this work, we approach robust dynamic obstacle avoidance twofold. First, we introduce the neuromorphic vision sensor (\textit{i.e.}, event camera) to provide \textbf{\textit{motion cues}} complementary to the traditional Laser depth data for handling dynamic obstacles. Second, we develop an DRL-based event-enhanced multimodal spiking actor network (EEM-SAN) that extracts information from motion events data via \textbf{\textit{unsupervised representation learning}} and fuses Laser and event camera data with \textbf{\textit{learnable thresholding}}. Experiments demonstrate that our EEM-SAN outperforms state-of-the-art obstacle avoidance methods by a significant margin, especially for dynamic obstacle avoidance.
\end{abstract}






\maketitle

\section{Introduction}
\label{sec:intro}
 \begin{figure*}
  \centering
  \includegraphics[width=1.0\textwidth]{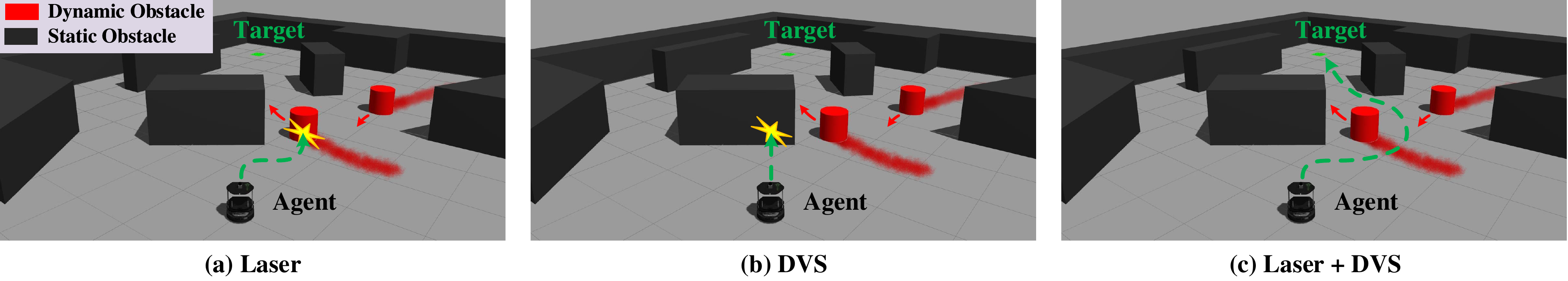} 
  \caption{Illustration of obstacle avoidance using different sensors: (a) Laser can help detect large static objects but fails to perceive fast-moving objects; (b) event camera DVS excels to capture moving objects but cannot provide depth information to avoid large textureless obstacles; and (c) combining Laser and DVS achieves robust obstacle avoidance.}
  \vspace{-6pt}
  \label{fig:motivation}
\end{figure*}

In robotics, obstacle avoidance is a fundamental yet challenging task of satisfying the control objective subject to non-intersection or non-collision position constraints. Autonomous obstacle avoidance is of vital importance for an intelligent agent such as a mobile robot to navigate in its environment and attracts more and more research attention in recent years \cite{montemerlo2002fastslam,pfeiffer2018reinforced,zhang2021learning,zhang2022closed} due to the growing need of practical usages such as post-disaster rescue \cite{2000RoboCup} and subterranean detection \cite{2020Complementary}. In order to be able to perform tasks in some high-risk or inaccessible scenarios instead of humans, the mobile robot needs to possess strong capabilities of robust and efficient autonomous navigation and dynamic obstacle avoidance.

A dependable autonomous navigation and obstacle avoidance algorithm, unlike the path planning method which involves the pre-computation of an obstacle-free path, needs to be implemented as a \textbf{\textit{reactive}} control law and should be \textbf{\textit{energy-efficient}} due to the limited computational resources on the mobile robots. As to the former, deep reinforcement learning (DRL) techniques \cite{mnih2015human} are widely used to train an agent to make good obstacle avoidance decisions in complex/unknown environments \cite{wang2018learning}. As to the latter, the biologically realistic spiking neural network (SNN) \cite{bohte2002error} in which neurons compute asynchronously and communicate through spikes is leveraged by existing state-of-the-art methods to achieve low power consumption and low latency obstacle avoidance \cite{dewolf2021spiking,abadia2021cerebellar,tang2020reinforcement}. With recent advances in deep learning, training SNN with RL for obstacle avoidance becomes the mainstream method in the field and achieves promising results. However, existing methods typically assume that the environment is static \cite{inproceedings,arakawa2020exploration} while in real-world scenes such as malls and streets, the obstacles (\eg, pedestrians and vehicles) are not always static but often \textbf{\textit{dynamic}}. The movement of obstacles makes the scene become more complex and leads to a faster relative speed between the agent and the obstacle, which requires the agent to make decisions and take avoidance actions in a shorter time, posing a great challenge to the existing methods. 
In this work, we strive to embrace challenges toward robust \textbf{\textit{dynamic obstacle avoidance}}. We approach this twofold. First, based on that the neuromorphic event camera called Dynamic Vision Sensor (DVS) \cite{2008A,gallego2020event} can record a \textit{high-frequency} stream of asynchronous brightness change events with extremely \textit{low latency} (in the order of $\mu$s) and \textit{low power consumption}, we for the first time introduce the two-dimensional DVS event modality into the SNN-based obstacle avoidance framework to provide motion cues complementary to the traditional Laser depth data for handling dynamic obstacles, building an omni perception of scenes, similar to human stereo vision \cite{marr1979computational}. Second, we develop an event-enhanced multimodal spiking actor network (EEM-SAN) that achieves dynamic obstacle avoidance in a deep reinforcement learning manner. EEM-SAN is built on three key modules: (i) a hybrid spiking variational autoencoder (HSVAE) that extracts information from DVS event data via unsupervised representation learning; (ii) a population coding (PC) module that combines population coding \cite{2020Deep} and Poisson coding \cite{tang2020reinforcement} to decode information from the activity of neurons; and (iii) a middle fuse decision module with learnable thresholding (MFDM-LT) designed for multimodal data fusion.


We perform extensive experiments to demonstrate the efficacy of our method and show that DVS events provide a powerful and complementary cue for dynamic obstacle avoidance (Figure \ref{fig:motivation}). In summary, our contributions are:

\begin{itemize}
\item the first solution to solve the challenging dynamic obstacle avoidance problem using a deep-reinforcement-learning-based spiking neural network with robot state and both Laser and DVS data as input, action decision as output;
\vspace{5pt}
\item a novel hybrid spiking variational autoencoder that decouples the representation learning of DVS event data from the whole reinforcement learning and greatly facilitates the training process;
\item a new middle fuse decision module with learnable thresholding to robustly integrate Laser and DVS data.
\end{itemize}

\section{Background and Related Work}
\label{sec_related}

\paragraph{Leaky Integrate-and-Fire (LIF) spiking neuron model} 
A spiking neural network (SNN) is a bio-plausible neural network that simulates biological information processing: neurons exchange information through spikes (or action potentials). Many different mathematical spiking neuron models have been developed, spanning from the simplest Integrate-and-Fire (IF) \cite{rathi2020diet} to sophisticated Spike Response Model (SRM) \cite{gerstner1995time}. Our approach leverages the \emph{Leaky Integrated-and-Fire (LIF) model} \cite{gerstner2002spiking}, a simplified variant of SRM model. We can define an $n_l$-layer feedforward SNN architecture with LIF neurons. Given $N^l$ incoming spike trains at layer $l$, $s_i^l(t)$, the SNN forward propagation is mathematically defined as:
\begin{align}
v_{i}^{l+1}(t) &=\sum_{j=1}^{N^{l}} w_{i j} s_{j}^{l}(t)+ \nonumber\\ 
&v_{i}^{l+1}(t-1) f_{d}\left(s_{i}^{l+1}(t-1)\right)+b_{i}^{l+1}, \nonumber\\
s_{i}^{l+1}(t) &=f_{s}\left(v_{i}^{l+1}(t)\right), \nonumber\\
f_{d}(s(t)) &= \begin{cases}D & s(t)=0 \label{eq:LIF} \\
0 & s(t)=1 \end{cases} ,
\end{align}
where $w_{ij}$ is the synaptic weight between the $j$-th neuron on the $l$-th layer and the $i$-th neuron on the layer $l+1$; $b_i^{l+1}$ is an adjustable bias, and $D$ is a constant. The operator $f_s(\cdot)$ is a spike function defined as:
\begin{align}
f_s(v) &: v \rightarrow s, s(t):= s(t) + \delta(t-t^{(f+1)}), \label{eq:srm_fs}\\
t^{f+1} &= \min\{t: v(t) = \Theta, t > t^{(f)}\}, \label{eq:srm_TH}
\end{align}
where $s(t)$ is a sequence of spikes called a spike train, $\delta$ is a mathematical function whose value is zero everywhere except at zero and whose integral over the entire real line is equal to one, and $\Theta$ is the membrane potential threshold which is static and the same for all neurons in the network \cite{tal1997computing}.

\paragraph{Dynamic Vision Sensor} DVS is a bio-inspired sensor that reports per-pixel brightness changes in log scale as a stream of asynchronous events \cite{gallego2020event,mei2023deep}. Compared to conventional frame-based cameras, DVS offers a very high dynamic range (140 dB versus 60 dB) and high temporal resolution (in the order of $\mu$s). An event, $e$, encodes three pieces of information: the pixel location, $(x, y)$, of an event, the timestamp, $t$, records the time when the event is triggered, and the polarity, $p \in \{-1, 1\}$, of an event, which reflects the direction of the changes. Formally, a set of events can be defined as
\begin{equation} 
\label{eq:event_set}
\mathcal{E} = \{e_k\}_{k=1}^N = \{[x_k, y_k, t_k, p_k]\}_{k=1}^N.
\end{equation}

In constant lighting conditions, events are triggered by moving edges (\eg, object contour and texture boundaries), making the DVS a natural edge extractor. However, this attractive feature also poses a unique challenge since events predominantly stem from edges, making the measured events inherently sparse. Asynchronous and sparse events cannot be effectively handled by CNN-based approaches designed for conventional frames.

\paragraph{DVS-based Robot Control}
Most robotics applications use traditional frame-based cameras as their perception devices. However, frame-based cameras have inherent characteristics such as high data volume, low temporal resolution, and high latency, which weakens their ability to perceive fast-moving objects and greatly limit the robot's manipulation capabilities. In this context, event cameras have attracted the attention of scholars in the field of robotics. By combining event cameras with robot perception and control, a series of breakthroughs in robot control has emerged, overcoming the limitations of traditional frame-based cameras. \cite{mueggler2015towards} is the first work to use DVS to avoid high-speed moving objects. To further explore the advantages of DVS, Falanga \etal \cite{falanga2019fast} conducted experiments on UAVs equipped with three sensors: monocular, stereo frame-based cameras, and DVS. The results showed that the DVS (2 $\sim$ 4 $ms$) had significantly lower delay than the monocular (26 $\sim$ 40 $ms$) and stereo frame-based cameras (17 $\sim$ 70 $ms$) when operating within the perception range of 2 $m$. Sanket \etal \cite{sanket2020evdodgenet} used a artificial neural network (ANN) to segment independent moving objects from event streams, and reasoned about their 3D motion to perform evasion tasks. 
Most of these methods rely on hand-crafted features or priors (\eg, Kalman filter~\cite{mueggler2015towards}, optical flow~\cite{falanga2019fast}, and the obstacles~\cite{sanket2020evdodgenet}) to perform obstacle reasoning and avoidance.
Our method differs from the above works in that we combine DRL and SNN with DVS to achieve robot control, enabling continuous autonomous robot navigation and robust yet efficient dynamic obstacle avoidance.

\paragraph{SNNs for Multimodal-Based Sensing}
SNNs have not been widely explored for multimodal-based sensing \cite{zhang2022spiking,zhang2023blink}. A few attempts have been made to combine image and audio modality. Liu \etal \cite{liu2022event} developed a weighted-sum-based attention scheme to fuse image and audio modalities. The weights for each modality are dynamically decided. In the same vein, Jia \etal \cite{jia2022motif} proposed an MR-SNN algorithm to fuse the same two modalities using a fusion mask. MR-SNN learns a Motif mask for each modality and generates a fusion mask by averaging the two learned masks. Instead of learning in a supervised manner, Rathi \etal \cite{rathi2018stdp} proposed an unsupervised multimodal learning method, which combines the image and audio modality by learning cross-modal connections enabled by the Spike Timing Dependent Plasticity (STDP) algorithm. All these methods assess their effectivenesses with a simple MNIST classification task. We validate our multimodal-based SNN approach with a more complex practical problem (\ie, obstacle avoidance), under both static and dynamic conditions. 

\paragraph{SNNs for Robot Control}
SNNs get substantial attention from the robot control community due to their high energy efficiency when deployed on neuromorphic hardware. A thread of work leverages SNNs for robot obstacle avoidance tasks \cite{inproceedings,mahadevuni2017navigating,tang2020reinforcement}. Tang \etal \cite{tang2020reinforcement} proposed a hybrid framework SDDPG for mapless navigation tasks. The SDDPG framework consists of an SNN-based actor network and a CNN-based critic network, where the two networks are co-trained together. Ding \etal \cite{ding2022biologically} proposed a bio-inspired dynamic spiking threshold scheme to enhance SDDPG's homeostasis in obstacle avoidance tasks under normal and degraded conditions. Later, Tang \etal \cite{2020Deep} extended their approach to continuous control tasks and proposed the PopSAN method. An essential contribution introduced by PopSAN is the population-coding scheme, which effectively addresses the high-dimensional state problem presented in continuous control tasks. Recently, Zhang \etal \cite{zhang2022multiscale} combined multiscale dynamic neurons coding and population coding to improve the performance of a spiking actor network. A critical difference between our approach and these methods is that we leverage two modalities for robot control instead of one.

\section{Methodology}
\label{ssec:overview}

\begin{figure*}[t]
  \centering
  \includegraphics[width=1\textwidth,height=14cm]{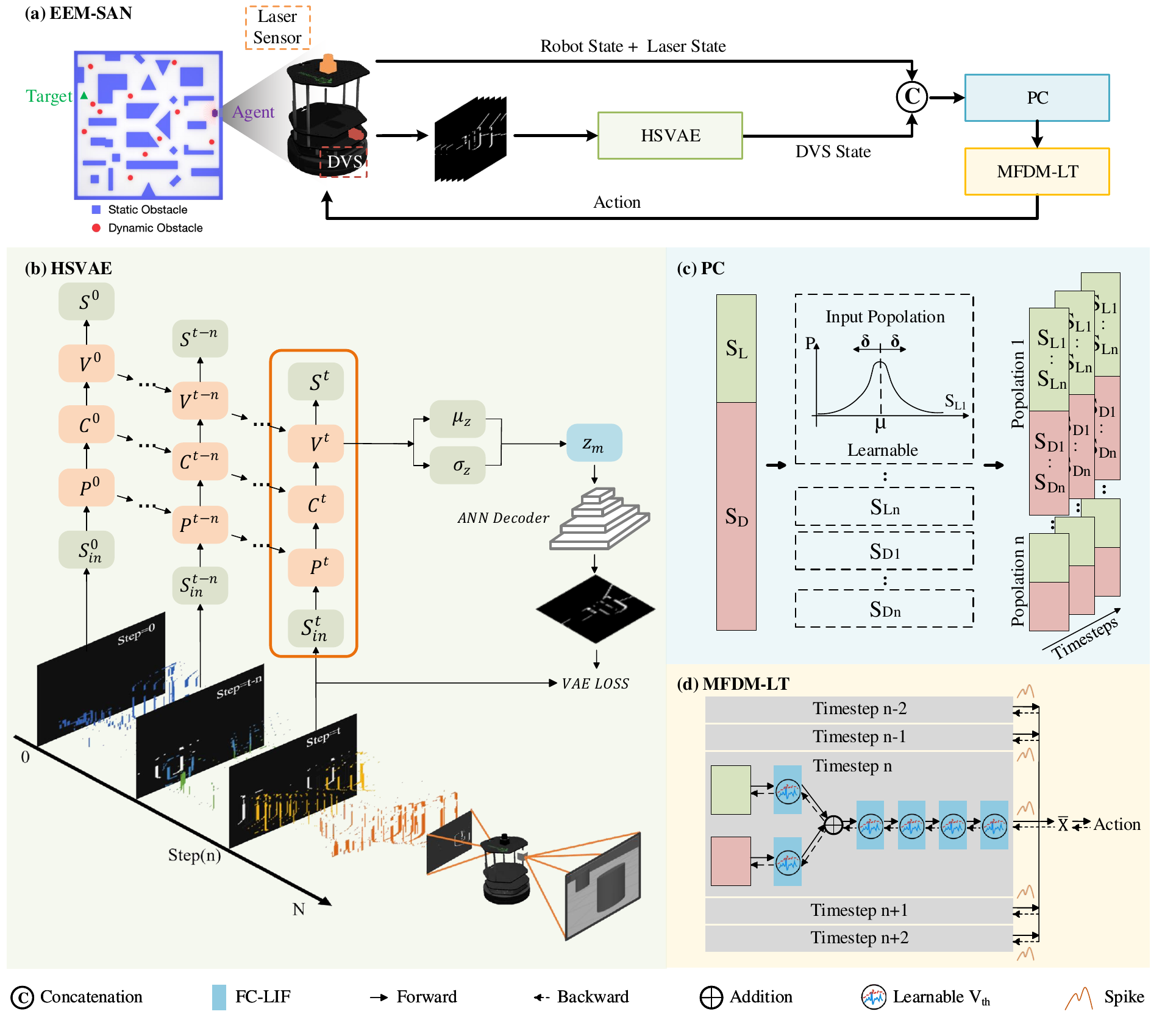} 
  \caption{Overview of EEM-SAN (a) and its three main components: (b) a Hybrid Spiking Variational Autoencoder (HSVAE) module, (c) a Population Coding (PC) module, and (d) a Middle Fuse Decision Module with Learnable Thresholding (MFDM-LT).} 
  \label{fig:pipeline}
  \vspace{-8pt}
\end{figure*}

\mhy{Figure \ref{fig:pipeline}(a) illustrates the diagram of our proposed event-enhanced multimodal spiking actor network (EEM-SAN) that is trained with reinforcement learning. First, the DVS sensor mounted on the agent records a high-frequency stream of brightness change events which is then encoded by a hybrid spiking variational autoencoder (HSVAE) (Figure \ref{fig:pipeline}(b)) into a one-dimensional vector (\ie, DVS State). Then, the obtained DVS state, together with the robot state and Laser state, is processed by a population coding (PC) module (Figure \ref{fig:pipeline}(c)) that can decode information from the activity of neurons. Finally, a middle fuse decision module with learnable thresholding (MFDM-LT) (Figure \ref{fig:pipeline}(d)) is used to fuse multimodal data and output obstacle avoidance decision which will be decoded into actual control action by Algorithm \ref{alg:ebm-san}.}

\begin{algorithm}[t] 
	\caption{Forward propagation through EEM-SAN}     
	 \label{alg:ebm-san}       
	\begin{algorithmic} 
	\Ensure  Left and right wheel speeds $v_{L}$, $v_{R}$
	\Require Min and max wheel speeds $v_{min}$, $v_{max}$
	\Require Robot State, Laser State, Event frame Deque:\\ $D_{E}$ = $(E_{1}, \cdots, E_{5})$
	\Require Learnable encoding means $\pmb{\mu}$ and standard deviations $\pmb{\sigma}$ for all populations
	\Require Middle Fuse Module: $MF$ and total timestep T
    \State DVS state generated by the Spiking Encoder of HSVAE: \textbf{DVS State} = HSVAE ($D_{E}$)
    \State Input Multimodal Observation $O$ = [\textbf{Robot State}, \textbf{Laser State}, \textbf{DVS State}]
    \State Spikes generated by the population coding module: \\ $\pmb{X}$ = PC ($\pmb{o}$, $\pmb{\mu}$, $\pmb{\sigma}$)
    \For {$t = 1, \cdots, T$}
        \State Laser spikes from populations at timestep $t$: \par $\pmb{S}_{L}^{(t)(0)}$ = $\pmb{X}^{(t)(O_{0}\cdots O_{L})}$
        \State DVS spikes from populations at timestep $t$: \par $\pmb{S}_{D}^{(t)(0)}$ = $\pmb{X}^{(t)(O_{D}\cdots O_{end})}$
        \State Fuse spikes from Middle Fuse Module:  \par $\pmb{S}_{M}^{(t)(0)}$ = $MF(\pmb{S}_{L}^{(t)(0)}, \pmb{S}_{D}^{(t)(0)})$
        \For {$k = 1, \cdots, K$}
        \State Update LIF neurons in layer $K$ at timestep $t$ based \par on spikes from layer $k$-1:
        \State $\pmb{c}^{(t)(k)}$ = $d_{c}\cdot \pmb{c}^{(t-1)(k)} + \pmb{W}^{(k)}\pmb{s}^{(t)(k-1)} + \pmb{b}^{(k)}$;
        \State $\pmb{v}^{(t)(k)}$ = $d_{v}\cdot \pmb{v}^{(t-1)(k)}\cdot (1-\pmb{s}^{(t-1)(k)}) + \pmb{c}^{(t)(k)}$;
        \State $\pmb{s}^{(t)(k)}$ = $Threshold(\pmb{v}^{(t)(k)})$;
        \EndFor
        \State $\textbf{SpikeCount}^{t}$ = $\textbf{SpikeCount}^{t-1}$ + $\pmb{s}^{(t)(l)}$
    \EndFor
    \State $\textbf{Action}$ = $\textbf{SpikeCount}^{(T)}/T$
    \State $v_{L}$ = $\textbf{Action}[0] * (v_{max} - v_{min}) + v_{min}$
    \State $v_{R}$ = $\textbf{Action}[1] * (v_{max} - v_{min}) + v_{min}$
  \end{algorithmic} 

\end{algorithm}

\subsection{Hybrid Spiking Variational Autoencoder}
\label{sec:svae}
\mhy{To exploit useful motion cues embedded in DVS event data, a key problem is how to effectively and efficiently learn its feature representations. Variational Autoencoder (VAE) \cite{kingma2013auto} is a powerful tool to extract the high-level embeddings of the input data. It typically consists of an encoder and decoder networks and is trained by:}

\begin{equation}
Loss=\frac{1}{w\times h}\sum_{i,j}^{w,h}(x_{i,j}-\overline{x_{i,j}})^{2} + \int p(x)log\frac{p(x)}{q(x)},
\end{equation}
where the first item is the mean squared error (MSE) between the original input image $x$ and the reconstruction $\overline{x}$, and the second item measures how the latent vector $p(x)$ matches a unit Gaussian $q(x)$ by Kullback-Leibler (KL) divergence. W and h represent the width and height of the event frame, respectively. 
%
VAEs have stable learning abilities in generative models and can be applied to various tasks. Recently, a series of works have attempted to use SNNs to create VAEs and make validation on classification datasets \cite{stewart2021gesture,talafha2020biologically,skatchkovsky2021learning}.
To take full advantage of the event-based nature of the continuous DVS stream and its rich temporal features, we design a novel hybrid spiking variational autoencoder (HSVAE) that contains an SNN encoder and ANN decoder. The architecture details of HSVAE are shown in Table \ref{t1}.
Unlike traditional spiking neural networks that reset the membrane potential, our method records all SNN-related states during one episode of simulated environment interaction. We record the current value, membrane potential and spike value of every neuron during one episode for the complete backpropagation chain. During deployment, we only need to record the relevant states at the last moment. We verified that SNNs trained in this way can integrate more temporal information to make better decisions. A single SNN layer of our HSVAE is shown in Figure \ref{fig:pipeline}(b) where refractory states are omitted for clarity. As illustrated, a stream of events recorded by the event camera mounted on a mobile robot is fed into SNN that can encode the spatio-temporal features of the input data into a latent state $z$. In Figure \ref{fig:pipeline}(b), $P$ is pre-synaptic potential, $C$ is current value and $V$ is the membrane potential of the spiking neuron. The training process of HSVAE is presented in Algorithm \ref{alg:sencoder}. 

\begin{table}[htbp]
  \begin{center}
  \caption{Architecture of our Hybrid Spiking VAE (HSVAE).}
   \label{t1}
    \small
    \begin{threeparttable} 
    \begin{tabular}{|c|c|c||c|} 
      \hline
      \textbf{Layer} & \textbf{Kernel} & \textbf{Output} & \textbf{Layer Type}\\
      \hline 
      input& & 128$\times$128$\times$1 & DVS128\\  \cline{4-4} 
      1 & 16c4p1s2 & 64$\times$64$\times$16 & \\  
      2 & 32c4p1s2 & 32$\times$32$\times$32 &  SNN LIF Encoder  \\
      3 & 16c4p1s2 & 16$\times$16$\times$16 &    \\
      4 & 1c4p1s2  & 8$\times$8$\times$1 &   \\ \cline{4-4} 
      5 & - & 64 & Flatten()    \\
      6 & - & 64 & $\mu$ = FC($V^t$)  \\
      7 & - & 64 & $\sigma$ = FC($V^t$)   \\
      8 & - & 8$\times$8$\times$1 & UnFlatten()    \\\cline{4-4} 
      9 & 16c4p1s2 & 16$\times$16$\times$16 &    \\
      10 & 32c4p1s2 & 32$\times$32$\times$32 &  ANN Decoder  \\
      11 & 16c4p1s2 & 64$\times$64$\times$16 &   \\ \cline{4-4} 
      12 & 1c4p1s2  & 128$\times$128$\times$1 & Event Frame  \\
      \hline
    \end{tabular}
      \begin{tablenotes}
      \centering
        \footnotesize         
        \item Notation: XcYpZsS represents channel X convolution filters (YxY) with padding $Z$ and stride $S$.     
    \end{tablenotes} 
    \end{threeparttable} 
  \end{center}
\end{table}

\begin{algorithm} [htbp]
	\caption{Spiking Neural Network Encoder Training}
	 \label{alg:sencoder}       
	\begin{algorithmic}
    \State Initialize the memory $D$ to store event frame observations,set the episode number $T$, set $e\_num = 0$;
    \State Start the simulation environment;
    \While {$e\_num < T$}
        \State Randomly set the start and goal locations in our training environment;
        \While {$not\ collision$}
        \State Capture the Event frame $X_t$,store $X_t$ in $D_T$;
        \State Move with the linear and angular velocity ;
        \EndWhile
    \EndWhile
    \For {$epoch = 1,T$}
        \State Initialize the SNN state $C_i,V_i,S_i$ in every layer;
        \For {$X_t = 1,D_t$}
        \State Update $\theta_0$ through on the loss defined in formula (4) and update SNN state $C_i,V_i,S_i$ through STBP algorithm;
        \EndFor
    \EndFor
  \end{algorithmic} 
\end{algorithm}

\subsection{Population Coding}
\label{sec:pop}
Neurons in the brain often use population coding and it isn't easy to decode correct information from the activity of a single neuron \cite{sun2017rate}. Hence, some researchers have begun to use populations of neurons to encode information into spike trains fed into SNNs instead of simple frequency encoding (\eg, Poisson coding) \cite{2020Deep,zhang2022multiscale}. To encode each dimension of the state, we created a population with 10 neurons, where each neuron has a Gaussian receptive field with two parameters: mean and standard deviation. These parameters were learned with surrogate backpropagation. Since SAN adopts the Poisson encoding, we apply the group encoding method combined with Poisson coding as PopSAN method in our experimental comparison. For a fair comparison, our method takes the same population and Poisson coding. Formally, the population coding function can be formulated as:
\begin{equation}
\left\{
\begin{aligned}
&A_{P_{i,j}}  =  EXP(-1/2\cdot ((s_{i}-\mu_{i,j})/\sigma_{i,j} )^{2}) \\
&\textbf{A}_{P} =  [A_{P_{1,1}}, \cdots,A_{P_{i,j}}, \cdots, A_{P_{N,J}}] 
\end{aligned},
\right.
\end{equation}
\begin{equation}
P(O_{k,t}=1) = C_{R}^{r}A_{P_{k}}^{r}(1-A_{P_{k}})^{R-r},
\end{equation}
where $i$ is the index of the input state ($i$ = 1, ..., N), $j$ is the index of neurons in a population (j = 1, ..., $J$), and $A_P$ is the stimulation strength after population coding, used for drawing the binary random number \cite{zhang2022multiscale}.

\subsection{Middle Fuse Decision Module with Learnable Thresholding}
The middle fuse decision module consists of a middle fuse (MF) module and a decision module (DM). 
\mhy{In MF, two modalities are first transformed to a one-dimensional vector with a length of 20 by two fully connected layers composed of LIF neurons, respectively, and the two obtained one-dimensional vectors are then fused via the element-wise addition.}
\mhy{DM contains four fully connected SNN layers and its forward propagation process is presented in Algorithm \ref{alg:ebm-san}.}
Since the threshold function of SNNs is non-differentiable, many methods focus on how to learn and tune it \cite{cheng2020finite,sporea2013supervised,zeng2017improving}. Among them, the approximate backpropagation method is widely due to its efficiency and flexibility. 
\mhy{Here, we adopt the STBP \cite{wu2018spatio} algorithm which uses
the rectangular function to approximate the gradient of a spike as follows:}
\begin{equation} 
z(V) =
\begin{cases}
1 & \text{if $\lvert V - V_{th} \rvert$ $<$ 0.5} \\
0 & \text{otherwise}\label{eq:stbp}  
\end{cases} ,
\end{equation}
where $z$ is the pseudo-gradient, $V$ is membrane potential, and $V_{th}$ is the threshold.

Several dynamic thresholding schemes have been developed in recent years to make neuronal thresholds learnable and varied throughout the network. However, most existing methods require strict prerequisites or are not based on the gradient descent method, which results in a higher computational load or makes it difficult to migrate to other SNNs \cite{meng2022training,kim2021spiking}. A few attempts have been made to integrate with the existing back-propagation-based training algorithms on classification tasks \cite{wang2022ltmd,yaoglif}.
Following the path, we further explore the performance of learnable thresholds under multimodal reinforcement learning tasks. We use the learnable thresholding mechanism in our MFDM. This mechanism is endowed with parameter optimization capability through STBP. By doing so, all neurons have different thresholds in the same network. This means the neuron’s response depends not only on its internal state but also on the threshold level. 
The key idea of the learnable thresholding is to find the comprehensive gradient of the loss function, and then the weight $W$ and the neuron’s threshold $H$ are simultaneously updated until convergence. From Eq. \ref{eq:LIF} and \ref{eq:stbp}, the threshold's partial derivatives of the loss function can be calculated as follows:
\begin{equation}
    \begin{aligned}
        \frac{\partial L}{\partial H^{n}} & = \sum_{t=1}^{T}\frac{\partial L}{\partial s^{t,n}}\frac{\partial s^{t,n}}{\partial H^{n}.} \\
        \label{eq:partial}
    \end{aligned}
\end{equation}

To ensure that the threshold remains within the appropriate area, we create a new parameter $r$ to define $H$ using hyperbolic tangent relation, formulated as $H$ = tanh ($r$). With this, Eq. \ref{eq:partial} can be expressed as:
\begin{equation}
    \begin{aligned}
        \frac{\partial L}{\partial r^{n}} & = \sum_{t=1}^{T}\frac{\partial L}{\partial s^{t,n}}\frac{\partial s^{t,n}}{\partial H^{n}}\frac{\partial H^{n}}{\partial r^{n}} \\ & = -\sum_{t=1}^{T}\frac{\partial L}{\partial s^{t,n}}f^{'}(v^{t,n} - tanh(r^{n}))(1-tanh^{2}(r^{n})).
    \end{aligned}
\end{equation}

Improved in this way, the threshold $H$ of the neurons can be iteratively trained using the backpropagation method and will be in the range (-1, 1). In this context, we apply the learnable thresholding scheme to MFDM and let neurons in the same layer share the same threshold to reduce learnable parameters.

\section{Experiments And Evaluation}
\label{sec:exp}

We evaluate the obstacle avoidance capabilities of the proposed EEM-SAN using success rate (SR) as a metric. SR is the percentage of successful passes among 200 trials. A successful pass is a trial in which a robot can reach the destination without touching any static or dynamic obstacles within 1000 steps. Our evaluation baseline model and test environment are modified variants of the SAN \cite{tang2020reinforcement} and its original simulated test environment, respectively. 

\subsection{Experimental Settings}
\mhy{\textbf{Implementation Details.}}
We integrated the EEM-SAN with the deep reinforcement learning algorithm DDPG \cite{lillicrap2015continuous}. We repeated each experiment three times to obtain the mean success rate. The frequency of the Laser sensor and DVS are set to 20 Hz and 100 Hz, respectively. The timestep of MFDM-LT is set as 5. The number of groups in the population coding is set as 10. The neuron current decay constant and the voltage decay constant are set as 0.5 and 0.75, respectively. During the training process, we set the batch size to 256 and the learning rate to 0.0001. 

\noindent\mhy{\textbf{Simulator Setup.}}
Our experiments are based on the Gazebo simulator \cite{koenig2004design}. Both the training and testing use the robot operating system (ROS) as a middleware. The testing environment was set to be different from the training environment for better validating the generalization capability of a method. Following existing methods SAN \cite{tang2020reinforcement}, PopSAN \cite{2020Deep}, and BDETT \cite{ding2022biologically}, we developed and tested our method in the same environment and with the same random seeds. Our testing environments are set to be very challenging to better validate the robustness of an obstacle avoidance method.
The challenges of our testing environments are in the following aspects: (i) \textbf{densely and highly dynamic obstacles} (eleven dynamic obstacles, all set to higher than the maximum speed of the robot); (ii) \textbf{faster robot speed} (the maximum speed of the robot is twice that in SAN); and (iii) \textbf{narrower traversal passages and more densely organized static obstacles}. We have experimentally demonstrated in Table \ref{tab:2} that the existing SOTA method SAN \cite{tang2020reinforcement} has a significant accuracy drop (\textit{i.e.}, from 97.8\% to 58.0\%) when transferred from common scenes to our challenging scenes. More details about training and testing environment can be found in Appendix and SAN~\cite{tang2020reinforcement}.


\subsection{Evaluation}
\begin{figure*}[htbp]
  \centering
  \includegraphics[width=1.0\textwidth]{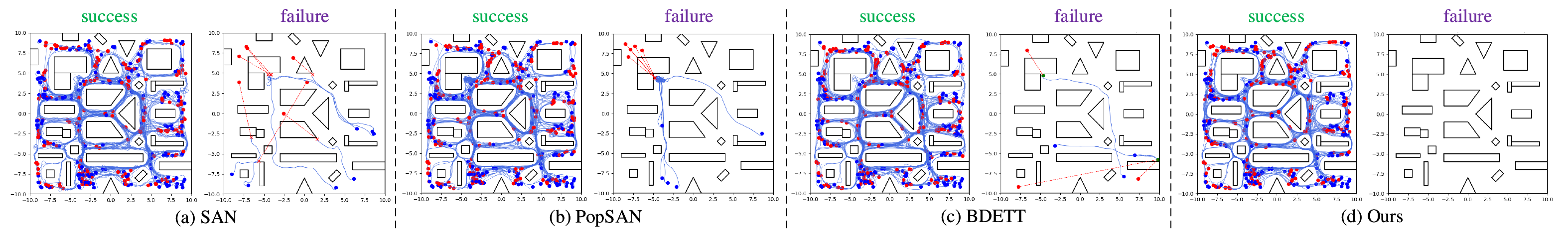} 

  \caption{Success and failure trajectories of the four methods at the same 200 randomly sampled start and goal locations. The blue dot represents the starting point, the red dot represents the target point, and the green dot represents the overtime case.} 
  \label{fig:3}
\end{figure*}

We extensively compare the effectiveness of our proposed EEM-SAN to three state-of-the-art methods including SAN \cite{tang2020reinforcement}, PopSAN \cite{2020Deep}, and BDETT \cite{ding2022biologically} across two different test maps \cite{tang2020reinforcement,ding2022biologically} under both dynamic and static conditions. The experimental results are reported in Table \ref{tab:2} and Figure \ref{fig:3}.

\paragraph{Dynamic Conditions}
To ensure the diversity and challenge of the scene, we set up eleven dynamic obstacles that reciprocate linearly along different trajectories at different speeds, and the speed of all moving obstacles is set to slightly higher than the maximum speed of the robot. From the results in Table \ref{tab:2}, we can see that our EEM-SAN outperforms all competing methods by a significant margin. For example, compared with the state-of-the-art method BDETT \cite{ding2022biologically}, our method improves $SR$ by $10.8\%$ and $11.8\%$ on the two testing maps, respectively. This clearly demonstrates the effectiveness of our method for dynamic obstacle avoidance. 

\paragraph{Static Conditions}
Although our method was originally designed for dynamic obstacle avoidance, we also tested its performance in static conditions to see its robustness. From Table \ref{tab:2}, we observe that: (1) our method still performs the best among all the compared methods for static obstacle avoidance; (2) when varying the maximum robot speed from 1.0 m/s to 0.5 m/s, all the methods get significant performance improvement. This is because a robot with a slow-moving speed has more time to make a decision and take action, which greatly decreases the difficulty of obstacle avoidance. Under such a condition, our method achieves very robust avoidance, \ie, the success rate $SR$ is up to 1.000; (3) compared to SAN \cite{tang2020reinforcement}, the PopSAN \cite{2020Deep} which is based on SAN \cite{tang2020reinforcement} but equipped with population coding method instead of frequency coding performs better, especially for the robot with a maximum speed of 1.0 m/s. This indicates that the population coding method is more suitable for handling rapidly changing scenes than the frequency coding method; and (4) by comparing the results of BDETT \cite{ding2022biologically} and PopSAN \cite{2020Deep}, the obvious performance improvement can be found under dynamic conditions but not under static conditions. This reveals that the dynamic thresholding scheme developed in BDETT \cite{ding2022biologically} benefits much from the dynamic scenes but little from static conditions. By contrast, our method can significantly improve obstacle avoidance performance in both dynamic and static conditions. Furthermore, Figure \ref{fig:3} visualizes an example that clearly shows that the existing methods often fail around the right-angle turn while our method can achieve robust obstacle avoidance.

\begin{table}[t]
	\centering
	\footnotesize
	\setlength{\tabcolsep}{6pt}{
		\caption{Quantitative performance of obstacle avoidance in both dynamic and static conditions on two different maps.}
		\begin{tabular}{cccc}
			\toprule
			\multirow{5}{*}{Methods} & Dynamic Conditions & \multicolumn{2}{c}{Static Conditions}\\

                \cmidrule(r){2-2}
                \cmidrule(r){3-4}
			
            & \makecell[c]{\tiny{Max Robot Speed}\\\tiny{1.0 m/s}} & \makecell[c]{\tiny{Max Robot Speed}\\\tiny{1.0 m/s}} & \makecell[c]{\tiny{Max Robot Speed}\\\tiny{0.5 m/s}} \\

            \cmidrule(r){2-2}
            \cmidrule(r){3-3}
            \cmidrule(r){4-4}
            & \tiny{Map 1 SR$\uparrow$ / Map 2 SR$\uparrow$} & \tiny{Map 1 SR$\uparrow$ / Map 2 SR$\uparrow$} & \tiny{Map 1 SR$\uparrow$ / Map 2 SR$\uparrow$} \\
    
			\midrule
			SAN&0.580 / 0.577&0.645 / 0.560&0.978 / 0.966\\
   
			\midrule
			PopSAN&0.598 / 0.618&0.805 / 0.718&0.983 / 0.973\\
                \midrule
                BDETT& 0.657 / 0.625& 0.735 / 0.728 & 0.975 / 0.923 \\
                 \midrule
			Ours&\pmb{0.765} / \pmb{0.743}&\pmb{0.870} / \pmb{0.848} &\pmb{1.000} / \pmb{0.985} \\
			\bottomrule
			
		\end{tabular}
    
		\label{tab:2}}
  \vspace{-5pt}
\end{table}

		
			

\subsection{Ablation Study}
\begin{figure*}[htbp]
  \centering
  \includegraphics[width=1.0\textwidth,height=3.5cm]{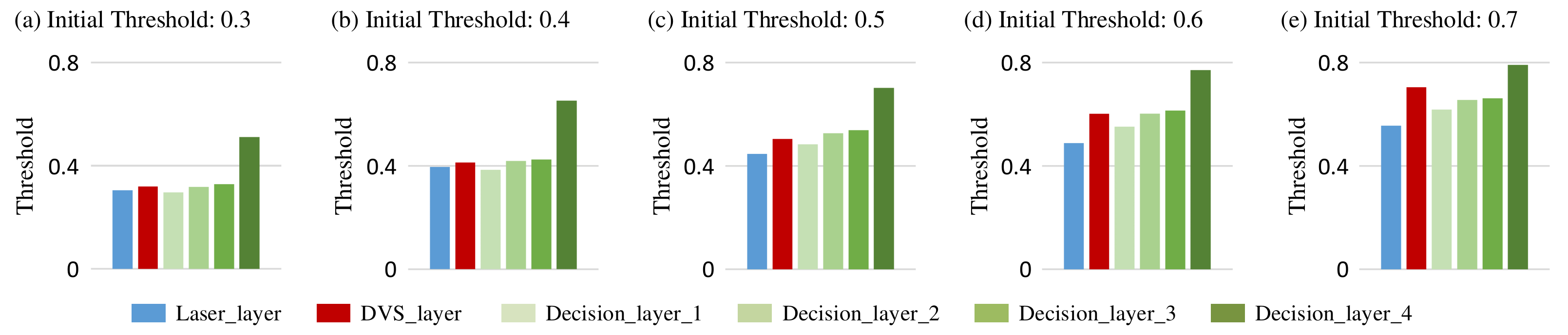} 
  \caption{Comparison between learned thresholds in different MFDM-LT layers under different initial training thresholds.} 
  \vspace{-13pt}
  \label{fig:4}
\end{figure*}

\begin{figure}[htbp]
  \centering
  \includegraphics[width=0.45\textwidth]{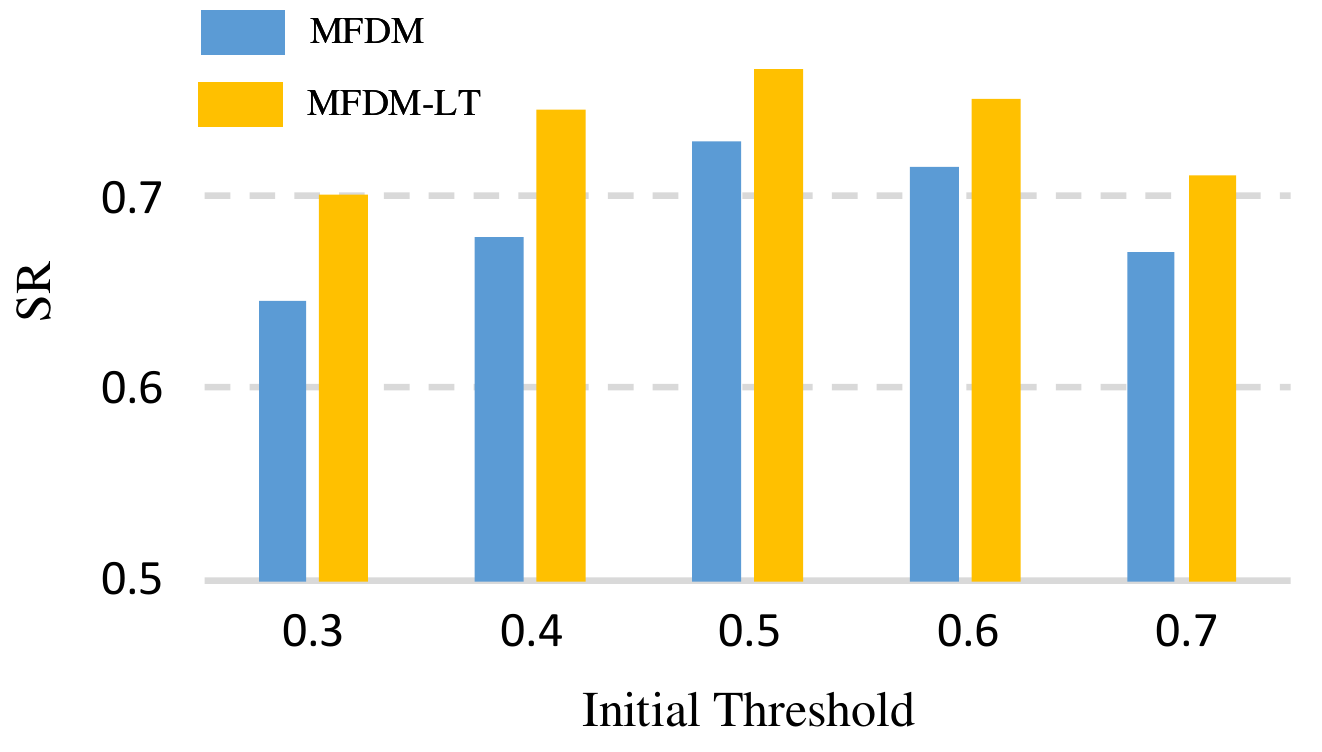} 
  \caption{Performance comparison between MFDM variants with and without learnable threshold scheme under different initial training thresholds settings.}
  \label{fig:5}
  \vspace{-13pt}
\end{figure}

To analyze our EEM-SAN, we investigate (a) the importance of DVS event cues; (b) the effectiveness of HSVAE; (c) the effectiveness of MFDM-LT; and (d) the efficiency advantage of SNN over ANN. Table \ref{tab:ab}, \ref{svae_compare}, \ref{computation_compare}, and Figure \ref{fig:4}, \ref{fig:5} summarize our findings.

\paragraph{Importance of DVS Event Cues}
We conduct an experiment to better understand the benefit of including DVS event cues for dynamic obstacle avoidance. Based on the baseline method PopSAN \cite{2020Deep} (Table \ref{tab:ab} (a)), we introduce the DVS event modality and implement an end-to-end (E2E) framework with naive modality fusion (\ie, addition). The $SR$ of (b) is much higher (\ie, 10.7\%) than that of (a), indicating DVS event is a strong cue for robust dynamic obstacle avoidance.



\begin{table}[t]
	\centering
    \footnotesize
	\setlength{\tabcolsep}{3mm}{
\caption{Quantitative ablation results indicate that each component in EEM-SAN contributes to the overall performance.}
		\begin{tabular}{c|cccc|c}
         \toprule
         \multirow{2}*{ }& Laser Only & \multicolumn{3}{|c|}{Laser + DVS Event}&\multirow{2}*{ SR $\uparrow$ }\\
            \cmidrule(r){2-5}
         & PopSAN & \multicolumn{1}{|c}{E2E} & HSVAE & MFDM-LT &\\
         \midrule
         (a)& $\surd$ & & & & 0.598 \\
         (b)& $\surd$ &$\surd$& & & 0.705 \\
         (c)& $\surd$ & &$\surd$& & 0.728 \\
         (d)& $\surd$ & &$\surd$ &$\surd$& \pmb{0.765} \\
        \bottomrule
        \end{tabular}
        
		\label{tab:ab}}
  \vspace{-5pt}
\end{table}

\paragraph{Effectiveness of HSVAE}
Based on ``E2E'' (Table \ref{tab:ab} (b)), adopting our proposed HSVAE (Table \ref{tab:ab} (c)) improves the $SR$ from 0.705 to 0.728. This demonstrates that our HSVAE can better extract information from DVS events frame for the decision making in obstacle avoidance. Furthermore, our HSVAE can be trained independently in advance and can be embedded in EEM-SAN with fixed weights during the reinforcement learning process. Without the need of saving/updating intermediate states like the DVS information extraction encoder in ``E2E'', our HSVAE can help save lots of GPU memory consumption and speed up the reinforcement learning process. Our HSVAE consists of an SNN encoder equipped with temporal information integration and an ANN decoder. It accommodates SNNs and ANNs in different layers, enabling the benefits of SNNs for sparse event data processing and ANNs for maintaining performance. 
The reason for using a hybrid architecture is that the spike activities decrease significantly as the network depth increases and the ANN decoder does not need to run during deployment. The experimental results in Table \ref{svae_compare} show that our HSVAE performs better than its ANN counterpart (AVAE) and the Fully Spiking Variational Autoencoder (FSVAE) \cite{kamata2022fully} whose encoder can not integrate temporal spikes information. This demonstrates that (i) the SNN encoder is more suitable for DVS information extraction than the ANN encoder; (ii) the temporal information integration is needed in the SNN encoder for the continuous obstacle avoidance decision; and (iii) the ANN decoder yields better results than the SNN decoder.

\begin{table}[t]
	\centering
    \footnotesize
	\setlength{\tabcolsep}{6mm}{
 \caption{Performance comparison between different DVS information encoding methods.}
		\begin{tabular}{cccc}
			\toprule
			Methods& AVAE & FSVAE &  HSVAE (ours)\\
			\midrule
            SR$\uparrow$& 0.677 &0.673 & \pmb{0.728}\\
			\bottomrule
			
		\end{tabular}
    
	\label{svae_compare}}
 \vspace{-5pt}
\end{table}

\begin{table}[t]
	\centering
         \footnotesize
         \renewcommand\arraystretch{0.7}
	\setlength{\tabcolsep}{3.5mm}{
  \caption{Comparison of the average amount of computation required to infer single state to action during one episode.}
		\begin{tabular}{c|c|c|c|c}
        	\toprule
        	 &\multicolumn{2}{c|}{Architecture} & \multicolumn{2}{c}{Computational Complexity}\\
        	\midrule
        	   & HSVAE& MFDM-LT& Addition & Multiplication \\
        	\midrule
                (i) & SNN&SNN& $\pmb{1.33 \times 10^{7}}$ & $\pmb{0.52 \times 10^{7}}$ \\
               \midrule
                (ii) & ANN&SNN &$5.87 \times 10^{7}$ & $5.78 \times 10^{7}$\\
                \midrule
                (iii) & SNN&ANN &$1.63 \times 10^{7}$ & $0.85 \times 10^{7}$\\
                \midrule
                (iv) & ANN&ANN & $6.16 \times 10^{7}$ & $6.11 \times 10^{7}$ \\
        	\bottomrule
        	\end{tabular}
        \vspace{-10pt}
		\label{computation_compare}}
\end{table}

\paragraph{Effectiveness of MFDM-LT}
Based on Table \ref{tab:ab} (c), replacing the naive addition modality fusion with our designed MFDM-LT (Table \ref{tab:ab} (d)) can greatly enhance the performance. To explore how the MFDM-LT works, we visualize the learned threshold values of different layers in the MFDM-LT module under different initial training threshold settings in Figure \ref{fig:4}. We can see that the learned thresholds of Laser and DVS layers are not equal. This shows that the Laser and DVS modalities contribute differently to the obstacle avoidance action. Besides, the learned threshold of DVS layer is higher than that of Laser layer for all the initial training threshold settings. The reason behind this may be (i) the DVS modality generates larger amount of data and thus a higher threshold is needed to filter out the useless spikes and (ii) lower threshold in Laser layer enables more depth information to pass through to dominate the action decision at the startup phase of the robot when lots of DVS events noise would be generated to interfere with decision making. Another interesting finding is that the threshold values in the decision layers increase as the network goes deeper. This phenomenon can be explained as the modality-fused data can fire neurons in shallow layers easily to maintain sufficient information which would be filtered by deep-layer neurons strictly to extract most informative data for the final action decision. Furthermore, we fix the learnable thresholds in MFDM-LT during the training process (Figure \ref{fig:5} MFDM) and find that the performance degrades dramatically. And this can be a strong evidence to demonstrate the effectiveness of the learnable threshold scheme in our MFDM-LT.

\paragraph{Efficiency Advantage of SNN over ANN}
Both HSVAE and MFDM-LT modules in our EEM-SAN are implemented by SNN, but not ANN. To show the advantage of such a choice, we compare the superior computational efficiency of our fully-SNN-based EEM-SAN against its ANN variants in terms of addition and multiplication FLOPs in Table \ref{computation_compare}. We can observe that, compared with the fully-SNN-based architecture (Table \ref{computation_compare} (i)), much more computations are required when adopting ANN-based HSVAE (ii), ANN-based MFDM-LT (iii), or fully-ANN-based network (iv). For example, the ANN version of EEM-SAN (iv) is 4.63 and 11.66 times more expensive than the SNN version (i) in terms of addition and multiplication operations, respectively. Prior works \cite{kim2022rate,yao2022attention,kamata2022fully} have demonstrated that the computing complexity of the network is positively correlated with inference speed and energy consumption, especially when the network is implemented in neuromorphic device~\cite{tang2020reinforcement,schoepe2021finding,blum2017neuromorphic}. Therefore, our fully-SNN-based EEM-SAN can achieve much faster inference with much lower energy consumption than its ANN counterpart.
\vspace{-8pt}
\subsection{Limitations}
In this work, we make the first attempt to connect multi-sensor representation and fully-SNN-based DRL towards robust and efficient robot control in extreme navigation scenarios with both static and fast-moving dynamic obstacles. Testing our method on different robot platforms (\eg, unmanned aerial vehicles) with more real challenging scenes (\eg, subterranean) would be a promising future work but out of focus of this work. Besides, our EEM-SAN has only been tested in a realistic simulator and has not been implemented on a real robot due to the unavailability of neuromorphic hardware. To conduct such an engineering verification, we are actively seeking permission from the Intel Neuromorphic Research Community (INRC) to use the neuromorphic chip Loihi. We will perform such an interesting verification once the hardware becomes available.

\vspace{-8pt}

\section{Conclusion}
In this paper, we presented an event-enhanced multimodal spiking actor network (EEM-SAN) for autonomous navigation and dependable obstacle avoidance. Our solution is the first to introduce the Dynamic Vision Sensor (DVS) to provide motion cues that complement the traditional Laser depth data for handling dynamic obstacles. EEM-SAN consists of two main modules: a hybrid spiking variational autoencoder (HSVAE) which encodes the DVS event data through unsupervised representation learning, and a middle fuse decision module with learnable thresholding (MFDM-LT) designed for multimodal data fusion. Through extensive validation and ablation studies, we demonstrate the value of DVS event cues, as well as the effectiveness and robustness of our EEM-SAN. In the future, we plan to deploy our method on neuromorphic devices to maximize its advantages in terms of computational efficiency and energy consumption.
\vspace{-8pt}

\begin{acks}
This work was supported in part by National Key Research and Development Program of China (2022ZD0210500), the National Natural Science Foundation of China under Grants 61972067/U21A20491, and the Distinguished Young Scholars Funding of Dalian (No. 2022RJ01). Ziqi Wei was supported by the open funding of State Key Laboratory of Structural Analysis for Industrial Equipment.
\end{acks}

\bibliographystyle{ACM-Reference-Format}
\bibliography{main}

\appendix
In the appendix, we first present more evaluation results under both dynamic and static conditions in Section \ref{sec:more}, and then detail more experimental settings in Section \ref{sec:settings}.

\section{More Evaluation Results}\label{sec:more}
\begin{figure}[H]
  \centering
  \includegraphics[width=0.5\textwidth]{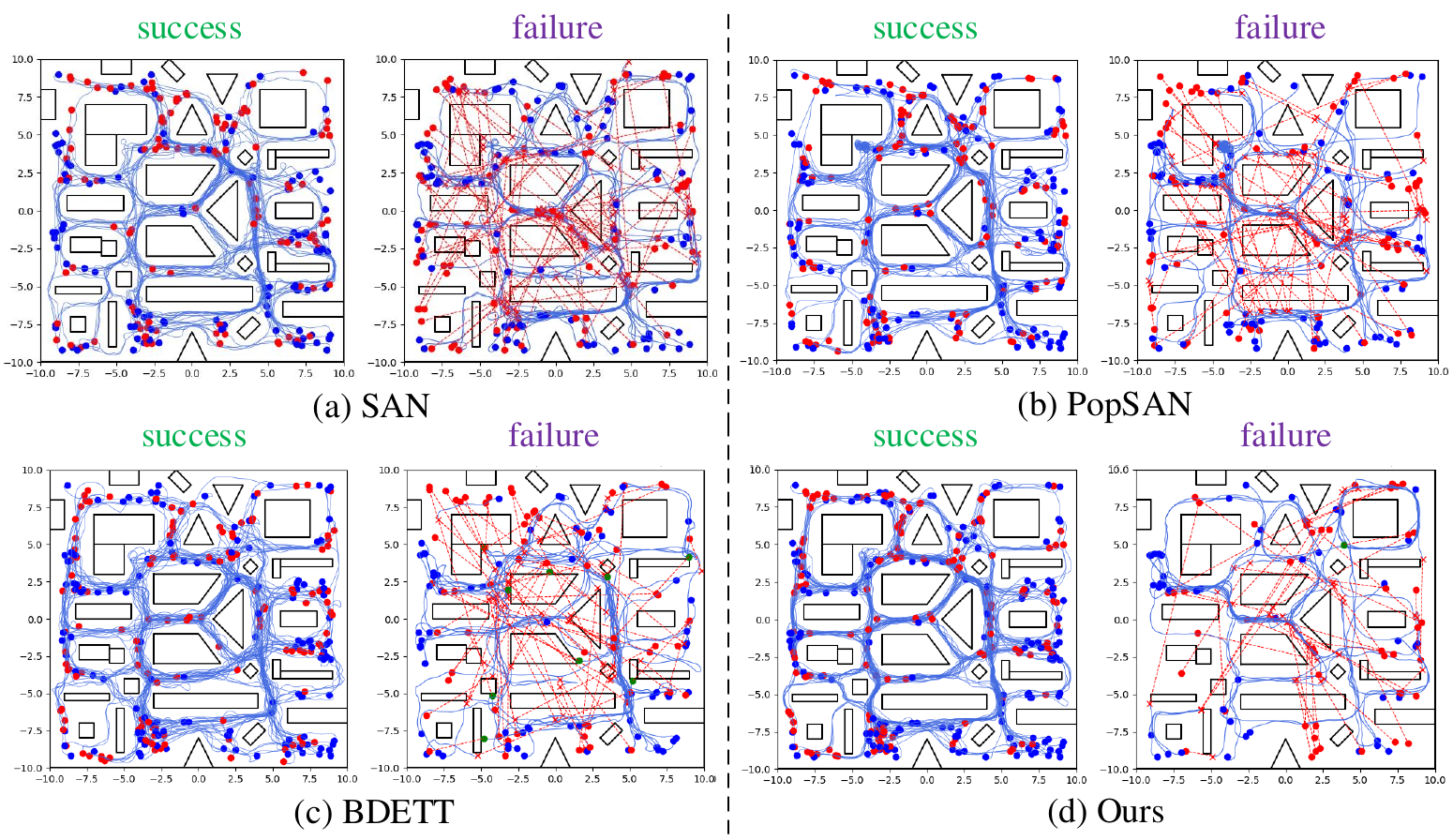} 
  \caption{Success and failure trajectories of the four methods at the same 200 randomly sampled start and goal locations \textit{in dynamic condition}. The blue dot represents the starting point, the red dot represents the target point, and the green dot represents the overtime case.} 
  \label{supfig:2}
\end{figure}

\begin{figure}[H]
  \centering
  \includegraphics[width=0.5\textwidth]{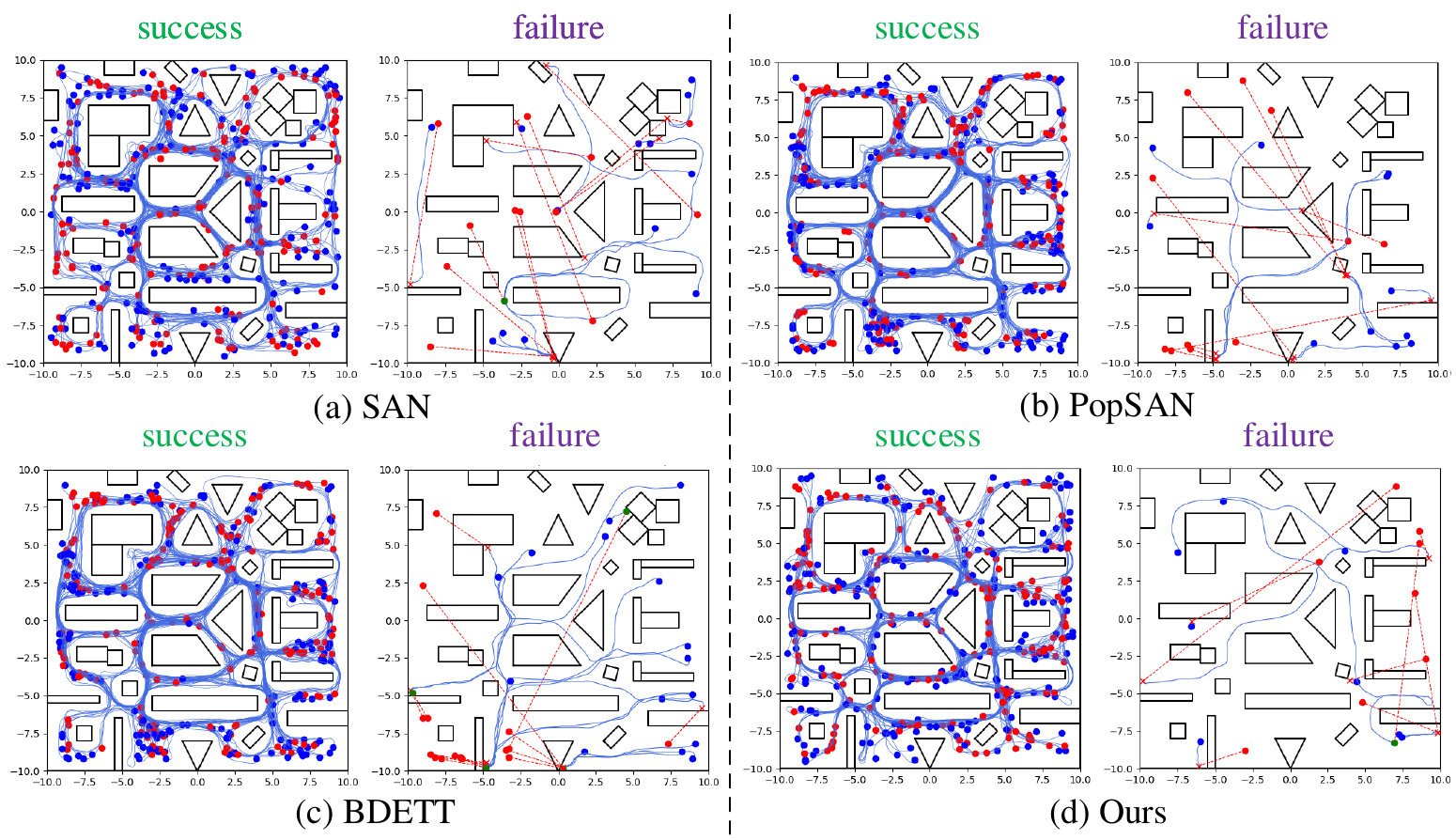} 
  
  \caption{Success and failure trajectories of the four methods at the same 200 randomly sampled start and goal locations \textit{in static condition}. The blue dot represents the starting point, the red dot represents the target point, and the green dot represents the overtime case.} 
  \label{supfig:1}
\end{figure}

Figure \ref{supfig:2} visualizes an example which clearly shows that the existing methods fail more times while our method can achieve robust obstacle avoidance in the dynamic condition. The comparison under the static condition is shown in Figure \ref{supfig:1} from which we can observe that our method still performs more robustly than the existing methods. 

\section{More Experimental Settings}\label{sec:settings}
\paragraph{Implementation Details.} 
The robot utilizes a Robo Peak LIght Detection And Ranging (RPLIDAR) system as its sensing device to detect obstacles, offering a field of view of 180 degrees with 18 range measurements. Each topic subscribed from gazebo contains 18-dimensional Laser data and 5 frames (128 $\times$ 128) of DVS data that are continuous in time. The event frame is generated from the DVS events stream through Algorithm \ref{alg:event_convert}. EEM-SAN was trained in conjunction with a deep critic network. More implementation details and hyperparameter configurations for our EEM-SAN were shown in Table \ref{list_hype}.

\begin{algorithm}[H]
    \caption{Converting events steam into event frame}     
    \label{alg:event_convert}       
    \begin{algorithmic}
        \Require a stream of event  $(e_t = (t,x,y,p))$,$t_n$,$W$,$H$   
        \State Initialize $s_n \gets O_{H,W}$
        \State Extract a subset of events $E \gets \{e_\gamma|t_{n-1} \le \gamma \le t_n\}$
        \State Initialize $s\_queue$ = $(5,d_n)$
        \While {event stream not stop}
            \For {$e_t \in E$}
                \State $t,x,y,p \gets e_t$
                \State $s\_n(x,y) \gets |p|$
                \State  $s\_queue.append(s_n)$
                \State  $s\_queue.popleft()$
            \EndFor
        \EndWhile
  \end{algorithmic} 
\end{algorithm} 

\begin{table}[H]
	\centering
	\setlength{\tabcolsep}{5mm}{
 \caption{Hyperparameter configurations.}
		\begin{tabular}{cc}
			\toprule
			Parameters &  Values\\
			\midrule
             Neurons per hidden layer for critic net & 512,512,512\\
             Actor learning rate & 1e-4\\
             Critic learning rate & 1e-4\\
             Reward discount factor & 0.99\\
             Maximum length of replay buffer & 1e5 \\
			\bottomrule
			
		\end{tabular}

	\label{list_hype}}
\end{table}

\begin{figure}[H]
  \centering
  \includegraphics[width=0.3\textwidth,height=5cm]{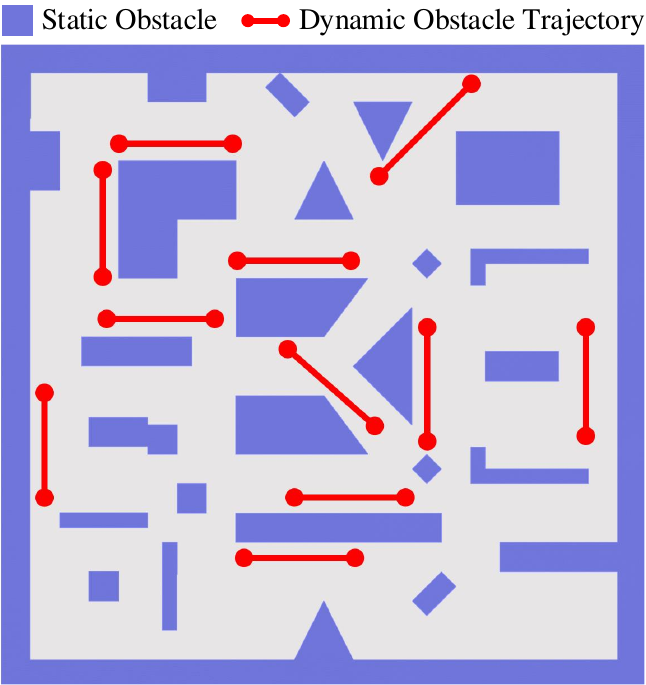} 

 \caption{A testing environment containing both static and dynamic obstacles, with blue representing static obstacles of different shapes and red lines representing the movement trajectory of dynamic obstacles.} 
  \label{supfig:0}
\end{figure}

\paragraph{Simulator Setup.} During our training phase, we utilized the same form of curriculum training method as in SAN \cite{tang2020reinforcement}, which has been proven to result in better generalization and faster convergence \cite{2017Reverse}. For our evaluation phase, we assessed our method in the Gazebo simulator \cite{koenig2004design} within a 20$m$ $\times$ 20$m$ test environment (Figure \ref{supfig:0}). The velocity of all moving obstacles within the test environment is listed in Table \ref{tab1}. To ensure comprehensive evaluation, we generated 200 start and goal locations, which were uniformly sampled at random from all parts of the test environment, with a minimum distance of 6$m$. We used the same start and goal locations to evaluate our method, as well as all the compared methods (\ie, SAN \cite{tang2020reinforcement}, PopSAN \cite{2020Deep}, and BDETT \cite{ding2022biologically}).

\begin{table}[H]
	\centering
    \footnotesize
	\setlength{\tabcolsep}{1.3mm}{
 \caption{The speed of all dynamic obstacles in the evaluation environment.}
		\begin{tabular}{|c|c|c|c|c|c|c|c|c|c|c|c|}
			\toprule
			Dynamic Obstacle & \# 1 & \# 2 & \# 3 & \# 4 & \# 5 & \# 6 & \# 7 & \# 8 & \# 9 & \# 10 &\# 11\\
			\midrule
            $V_x$ (m/s) & 1.4 & 1.4 & 1.4 & 1.2 & 0 & 0 & 0 & 1.4 & 1.3 & 1.4 & 0  \\
		  \midrule
		 $V_y$ (m/s) & 0 & 0 & 0 & 1.2 & 1.4 & 1.4 & 1.4 & 0 & 1.3 & 0 & 1.4\\
			\bottomrule
			
		\end{tabular}
    
	\label{tab1}}
\end{table}

\newpage

\end{document}